\documentclass{article}
\usepackage{spconf}
\usepackage{epsfig, graphicx}
\usepackage{amsmath}
\usepackage{amssymb}
\usepackage{cite}
\usepackage{subfigure}

\title{Two-Dimensional ARMA Modeling for Breast Cancer Detection and Classification}
\twoauthors
  {N. Bouaynaya, J. Zielinski
}
   {University of Arkansas at Little Rock\\
   Department of Systems Engineering}
  {D. Schonfeld}
   {University of Illinois at Chicago\\
   Department of Electrical and Computer Engineering}
\begin{document}
%\ninept
%
\maketitle
\begin{abstract}
We propose a new model-based computer-aided diagnosis (CAD) system
for tumor detection and classification (cancerous v.s. benign) in
breast images. Specifically, we show that (x-ray, ultrasound and
MRI) images can be accurately modeled by two-dimensional
autoregressive-moving average (ARMA) random fields. We derive a
two-stage Yule-Walker Least-Squares estimates of the model
parameters, which are subsequently used as the basis for statistical
inference and biophysical interpretation of the breast image. We use
a k-means classifier to segment the breast image into three regions:
healthy tissue, benign tumor, and cancerous tumor. Our simulation
results on ultrasound breast images illustrate the power of the
proposed approach.
\end{abstract}
\section{Introduction}
\label{sec:intro} Breast cancer continues to be a significant public
health problem in the United States: It is the second leading cause
of female mortality, and, disturbingly, one out of eight women in
the United States will be diagnosed with breast cancer in her life
time. Until the cause of this disease is fully understood, early
detection remains the only hope to improve breast cancer prognosis
and treatment. Breast cancer screening modalities are mainly based
on clinical examination, mammography, ultrasound imaging, magnetic
resonance imaging (MRI), and core biopsy. Mammography (breast x-ray
imaging) is by far the fastest and cheapest screening test for
breast cancer. Unfortunately, it is also among the most difficult of
radiological images to interpret: mammograms are of low contrast,
and features indicative of breast disease are often very small. Many
studies have shown that ultrasound and MRI imaging techniques can
help supplement mammography by detecting small breast cancers that
may not be visible with mammography. However, these techniques often
fail to determine if a detected tumor is cancerous or benign, and a
biopsy may be recommended. Consequently, many unnecessary biopsies
are
often undertaken due to the high false positive rate. %Moreover, high
%quality MRI images are assured only if the patient is able to remain
%perfectly still while images are being recorded. Otherwise, the
%patient's movements, due to anxiety and/or unease, will result in
%unclear images. Finally, many clinical facilities do not offer
%ultrasound or MRI screening, and the procedures are costly and are
%not covered by many insurance plans.

Computer aided diagnosis (CAD) paradigms have recently received
great attention for lesion detection and discrimination in X-ray and
ultrasound breast mammograms \cite{ElTE07, WYNJ05, MSLF08, PGWC08}.
The large amount of negative biopsies encountered in clinical
practice could be reduced if a computer system was available to help
the radiologists screen breast images. Broadly, the CAD systems
proposed in the literature can be grouped into four major
categories: geometrical \cite{ElTE07}, artificial intelligence
\cite{WYNJ05}, pyramidal (or multiresolution) \cite{MSLF08}, and
model-based techniques \cite{PGWC08}, \cite{AYBF04}. Geometrical
methods employ morphological and other segmentation techniques to
extract small specks of calcium known as microcalcifications from
breast images \cite{ElTE07}. However, this procedure usually
requires a priori knowledge of the tumor pattern characteristics.
Moreover, these techniques also tend to rely on many stages of
heuristics attempting to eliminate false positives. Artificial
intelligence techniques include neural networks and fuzzy logic
methods. The performance of these systems is tied to the
architecture of the network and the number of training data. Breast
cancer is a heterogeneous disease which includes several subtypes
with distinct prognosis. In particular, the variability associated
with the appearances of the breast cancer, ranging from relative
uniformity to complex patterns of bright streaks and blobs
\cite{WYNJ05}, makes the ANN require a large training data set to
ensure a certain level of reliability. Pyramidal or multiresolution
techniques refer mainly to the wavelet transform \cite{MSLF08},
which can be seen from a signal decomposition view point.
Specifically, a signal is decomposed onto a set of the basis wavelet
functions. A very appealing feature of the wavelet analysis is that
it provides a uniform resolution for all the scales. Limited by the
size of the basic wavelet function, the downside of the uniform
resolution is uniformly poor resolution. Model-based methods include
linear, non-linear and finite-element methods to build an accurate
model of the breast \cite{PGWC08}, \cite{AYBF04}. The model is
subsequently used for image matching, detection, and classification
\cite{AYBF04}. The accuracy of the results are tied to the accuracy
of the considered model.

In this work, we propose a new model-based CAD system for tumor
detection and classification. We show that (x-ray, ultrasound, and
MRI) breast images can be accurately modeled by two-dimensional
autoregressive moving average (ARMA) random fields. The model
parameters, being the fingerprints of the image, serve as the basis
for statistical inference and biophysical interpretation of the
breast image. ARMA models are parametric representations of
wide-sense stationary (WSS) processes with rational spectra.  The
Wold decomposition theorem states that any WSS process can be
decomposed as the sum of a regular process, which spectrum is
continuous, and a predictable process, which spectrum consists of
impulses. Since rational spectra form a dense set in the class of
continuous spectra, the ARMA model renders accurately the regular
part of the WSS process. It is, therefore, surprising that very few
researchers have attempted to derive a general ARMA representation
of the breast, and use it for tumor detection and classification. In
\cite{AYBF04}, the authors use a one-dimensional fractional
differencing autoregressive moving average (FARMA) process to model
the ultrasound RF echo reflected from the breast tissue. However, by
considering separate scan lines, they do not take into account the
two-dimensional spatial correlation between the pixels in the image.
In \cite{LaNo07}, an autoregressive (AR) model is considered for
improving the contrast of breast cancer lesions in ultrasound
images. ARMA models, however, provide a more accurate model of a
homogeneous random field than an AR model. As in the 1D case, the 2D
ARMA parameter estimation problem is much more difficult than its 2D
AR counterpart, due to the non-linearity in estimating the 2D moving
average (MA) parameters.

This paper is organized as follows: In Section \ref{sec:1}, we
define the 2D ARMA representation of the breast image, and derive a
Yule-Walker Least squares estimates of its parameters \cite{StMo05}.
In Section \ref{sec:2}, we use the estimated ARMA coefficients as
vector features for the k-means classifier. The simulation results,
for ultrasound breast images showing cancerous and benign tumors,
are shown in Section \ref{sec:3}. Finally, in Section
\ref{sec:conclusion}, we summarize our contribution and provide
concluding remarks.

\section{2D ARMA Modeling} \label{sec:1}
We represent the breast image as a 2D random field $\{x[n,m], \\
(n,m) \in \mathbb{Z}^2\}$. We define a total order on the discrete
lattice as follows
\begin{equation} \label{eq:order}
(i,j) \leq (s,t) \Longleftrightarrow i \leq s ~~\text{and}~~j\leq t.
\end{equation}
The 2D ARMA($p_1,p_2,q_1,q_2$) model is defined for the $N_1 \times
N_2$ image $I= \{x[n,m]: 0 \leq n \leq N_1-1, 0 \leq m \leq N_2-1\}$
by the following difference equation
\begin{eqnarray} \label{eq:model}
x[n,m] + \mathop{\sum_{i=0}^{p_1} \sum_{j=0}^{p_2}}_{(i,j) \neq (0,0)} a_{ij} x[n-i,m-j] = \nonumber \\
w[n,m] + \mathop{\sum_{i=0}^{q_1} \sum_{j=0}^{q_2}}_{{(i,j) \neq
(0,0)}} b_{ij} w[n-i,m-j],
\end{eqnarray}
where $\{w[n,m]\}$ is a white noise field with variance $\sigma^2$,
and the coefficients $\{a_{ij}\}, \{b_{ij}\}$ are the parameters of
the model. From Eq. (\ref{eq:model}), the image $\{x[n,m]\}$ can be
viewed as the output of the linear time-invariant causal system
$H(z_1,z_2)$ excited by a white noise input, where
\begin{equation} \label{eq:filter}
H(z_1,z_2) = \frac{B(z_1,z_2)}{A(z_1,z_2)} = \frac{\sum_{i=0}^{q_1}
\sum_{j=0}^{q_2} b_{ij}~ z_1^{-i} z_2^{-j}}{\sum_{i=0}^{p_1}
\sum_{j=0}^{p_2} a_{ij}~ z_1^{-i} z_2^{-j}},
\end{equation}
with $a_{00} = b_{00} = 1$. %The filtering referred to in Eq.
%(\ref{eq:filter}) can be written in the time domain as
%\begin{equation}
%A(z_1,z_2) x[n,m] = B(z_1,z_2) w[n,m].
%\end{equation}
%If $B(z_1,z_2) = 1$ (i.e, $q_1 = q_2 = 0$), then $x[n,m]$ is an
%autoregressive (AR) field; and $x[n,m]$ is a moving average (MA)
%field if $p_1 = p_2 = 0$.

\subsection{Yule-Walker Least-Squares Parameter Estimation}
Assume first that the noise sequence $\{w[n,m]\}$ were known. Then
the problem of estimating the parameters in the ARMA model
(\ref{eq:model}) would be a simple input-output system parameter
estimation problem, which could be solved by several methods, the
simplest of which is the least-squares (LS) method. In the LS
method, we express Eq. (\ref{eq:model}) as
\begin{equation} \label{eq:LS1}
x[n,m] + \phi^t[n,m] \theta = w[n,m],
\end{equation}
where
\begin{eqnarray*}
\phi^t[n,m] &=& [x[n,m-1], \cdots, x[n-p_1,m-p_2],-w[n,m-1], \nonumber \\
 && \cdots, -w[n-q_1,m-q_2]],
\end{eqnarray*}
and
\begin{displaymath}
\theta = [a_{01}, \cdots, a_{p_1p_2},b_{01}, \cdots, b_{q_1q_2}]^t.
\end{displaymath}
Writing Eq. (\ref{eq:LS1}) in matrix form for $n=L+1, \cdots,
N_1-1$, and $m = M+1, \cdots, N_2-1,$~ for some $L > \max(p_1,q_1)$,
and $M > \max(p_2, q_2)$, gives
\begin{equation} \label{eq:matrix}
\mathbf{x} + \Phi \mathbf{\theta} = \mathbf{w},
\end{equation}
where
\begin{eqnarray*}
&&\mathbf{x} = [x[L+1,M+1], \cdots, x[N_1-1,N_2-1]]^t, \\
&&\mathbf{w} = [w[L+1,M+1], \cdots, x[N_1-1,N_2-1]]^t,
\end{eqnarray*}
and $\Phi$ is displayed below.
\begin{figure*}
\begin{displaymath}
\Phi = \left(
\begin{array}{cccccc}
             x[L+1,M] & \cdots &  x[L+1-p_1,M+1-p_2] &  -w[L+1,M] &  \cdots &  -w[L+1-q_1,M+1-q_2] \\
             x[L+2,M] &  \cdots & x[L+2-p_1,M+1-p_2] &  -w[L+2,M] &  \cdots &  -w[L+2-q_1,M+1-q_2] \\
              \vdots & &  \vdots &   \vdots &   &   \vdots \\
             x[N_1-1,N_2-2] & \cdots &  x[N_1-1-p_1,N_2-1-p_2] & -w[N_1-1,N_2-2] &   \cdots &  -w[N_1-1-q_1,N_2-1-q_2] \\
            \end{array}
\right).
\end{displaymath}
\end{figure*}
Assume we know $\Phi$, then we can obtain a least-squares estimate
of the parameter vector $\theta$ in Eq. (\ref{eq:matrix}) as
\begin{equation} \label{eq:sol}
\hat{\theta} = - (\Phi^t \Phi)^{-1} \Phi^t \mathbf{x}.
\end{equation}
Observe that the input model noise $\{w[n,m]\}$ in $\Phi$ is
unknown. Nevertheless, it can be estimated by considering the noise
process $w[n,m]$ as the output of the linear filter
$\frac{1}{H(z_1,z_2)} = \frac{A(z_1,z_2)}{B(z_1,z_2)}$ with input
$x[n,m]$. From Nirenberg's proof of the division theorem in
multi-dimensional spaces \cite{Niren75}, we can write the inverse
ARMA filter $\frac{A(z_1,z_2)}{B(z_1,z_2)}$ as the infinite order AR
filter $ \sum_{i=0}^\infty \sum_{j=0}^\infty \alpha_{ij} z_1^{-i}
z_2^{-j}$. In the time domain, we obtain
\begin{equation} \label{eq:infinite-AR}
x[n,m] + \mathop{\sum_{i=0}^\infty \sum_{j=0}^\infty}_{(i,j) \neq
(0,0)} \alpha_{ij}~ x[n-i,m-j] = w[n,m].
\end{equation}
Therefore, we can estimate $\{w[n,m]\}$ by first estimating the AR
parameters $\{\alpha_{ij}\}$ and next obtaining $\{w[n,m]\}$ by
filtering $\{x[n,m]\}$ as in Eq. (\ref{eq:infinite-AR}). Since we
cannot estimate an infinite number of (independent) parameters from
a finite number of samples, we approximate the finite AR model by
one of finite order, say $(K_1,K_2)$. The parameters in the
truncated AR model can be estimated by using a 2D extension of the
Yule-Walker equations as follows
\begin{equation} \label{eq:YWAR}
r[k,l] + \mathop{\sum_{i=0}^{K_1} \sum_{j=0}^{K_2}}_{(i,j) \neq
(0,0)} \alpha_{ij} r[k-i,l-j] = \sigma^2 \delta[k,l],
\end{equation}
where $\{r[k,l]\}$ are the autocorrelation values of the field
$\{x[n,m]\}$, computed as follows
\begin{eqnarray}
&& r[k,l] = \frac{1}{(N-k)(M-l)} \sum_{i = 1}^{N-k} \sum_{j=1}^{M-l}
x[i,j] x[i+k,j+l], \nonumber \\
&& r[-k,-l] = r[k,l], ~\text{for}~ (k,l) \geq (0,0) \nonumber\\
&& r[k,-l] = r[-k,l], ~\text{for}~ (k,l) \geq (1,1),
\end{eqnarray}
and $\delta[k,l]$ is the 2D Kronecker delta function.
%\begin{displaymath}
%$\delta[k,l] = \left\{
%                 \begin{array}{ll}
%                   1, & \hbox{if $(k,l) = (0,0)$;} \\
%                   0, & \hbox{Otherwise.}
%                 \end{array}
%               \right.$
%\end{displaymath}
Equation (\ref{eq:YWAR}) is a system of linear equations that can be
written in matrix form and solved for the coefficients
$\alpha_{ij}$. Finally, the Yule-Walker Least-Squares algorithm is
summarized below
\begin{enumerate}
  \item Estimate the parameters $\{\alpha_{ij}\}$ in an AR($K_1,K_2$)
model of $x[n,m]$ by the YW method in (\ref{eq:YWAR}). Obtain an
estimate of the noise field $\{w[n,m]\}$ by
\begin{displaymath}
\hat{w}[n,m] = x[n,m] + \mathop{\sum_{i=0}^{K_1}
\sum_{j=0}^{K_2}}_{(i,j) \neq (0,0)} \hat{\alpha}_{ij} x[n-i,m-j],
\end{displaymath}
for $n = K_1+1, \cdots, N_1$, and $m = K_2+1,\cdots,N_2$.
  \item Replace the $w[n,m]$ by $\hat{w}[n,m]$ computed in Step 1.
Obtain $\hat{\theta}$ in (\ref{eq:sol}) with $L = K_1 + q_1$, and $M
= K_2 + q_2$.
\end{enumerate}

\section{Tumor Detection and Classification} \label{sec:2}
The estimated ARMA parameters, $\{a_{ij}\}, \{b_{ij}\}$, are used as
a basis for inference about the presence of a tumor and its nature:
benign or cancerous. We use the k-means algorithm to segment the
breast image into 3 classes: healthy tissue, benign tumor and
cancerous tumor. Our method consists of representing each pixel in
the image by an ARMA model whose parameters are estimated by using
an appropriate neighborhood for the pixel. We make the assumption
that all pixels in the considered neighborhood belong to the same
class, and hence, for computational efficiency, we replace the
entire neighborhood by the vector value of the estimated ARMA
parameters. This procedure is repeated for the entire image,
creating a new block by block vector-valued image, which will be the
input to the k-means classifier.

\begin{figure*}[t]
  \centering
  \includegraphics[width= 1.45in]{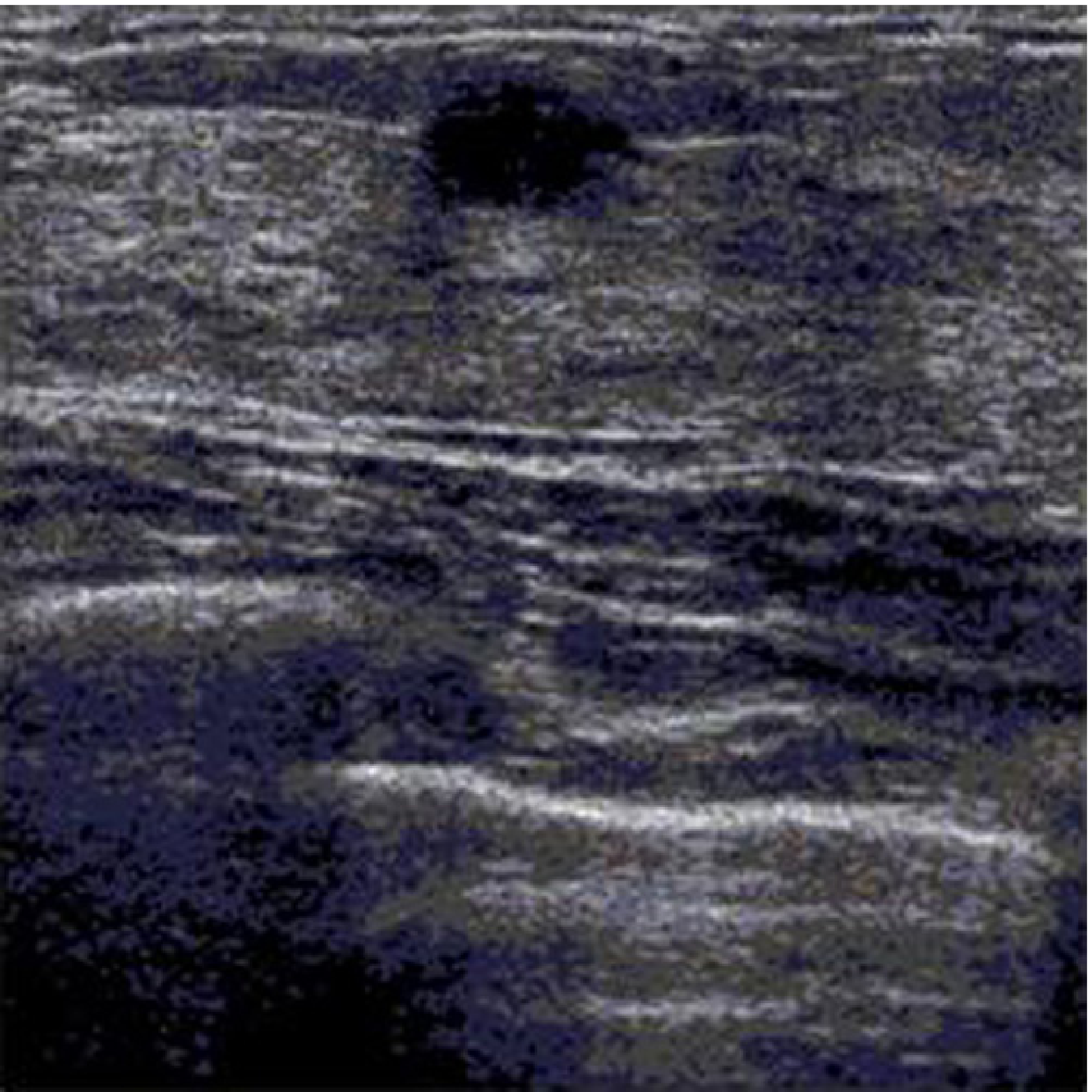}
\includegraphics[width= 1.45in]{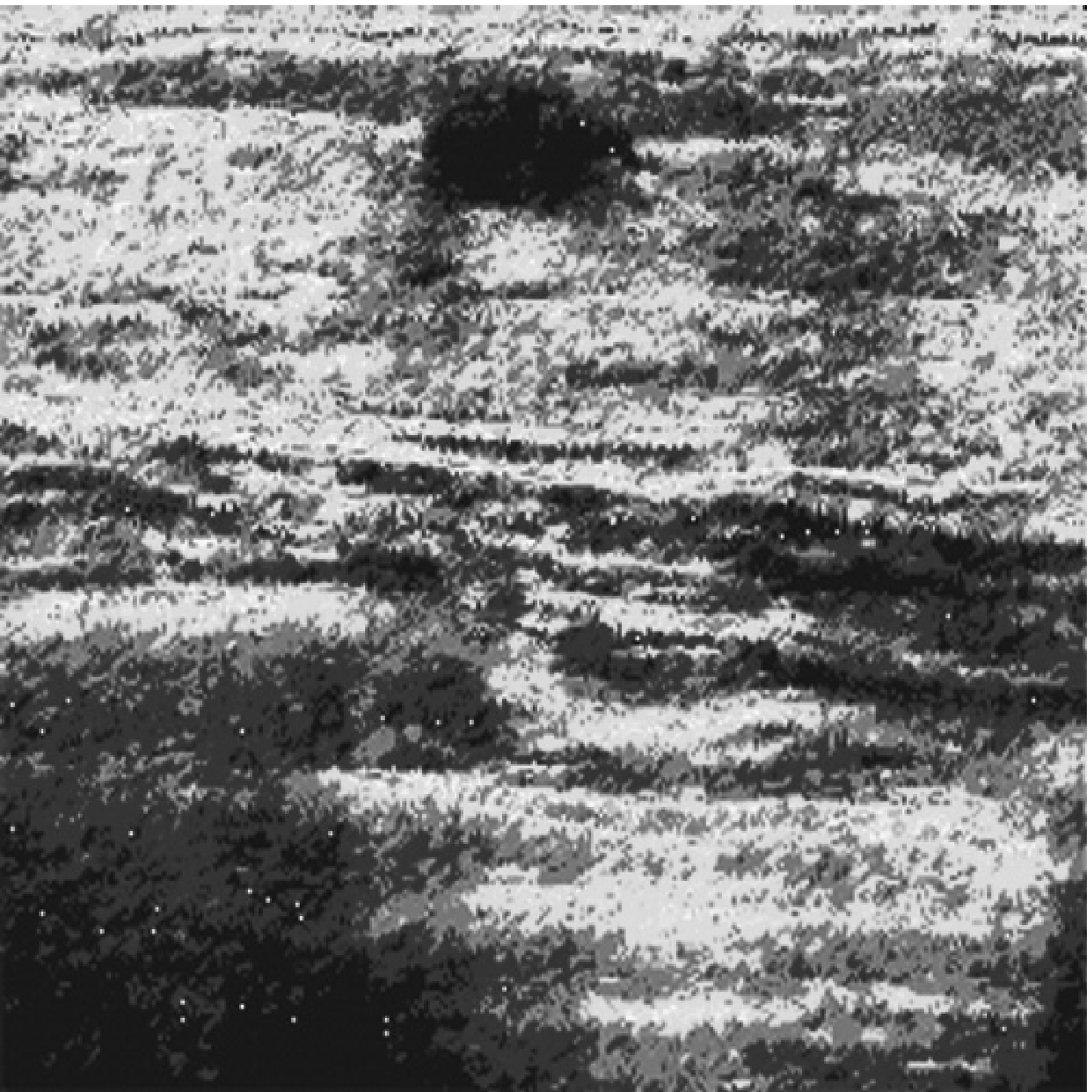}
\includegraphics[width= 1.45in]{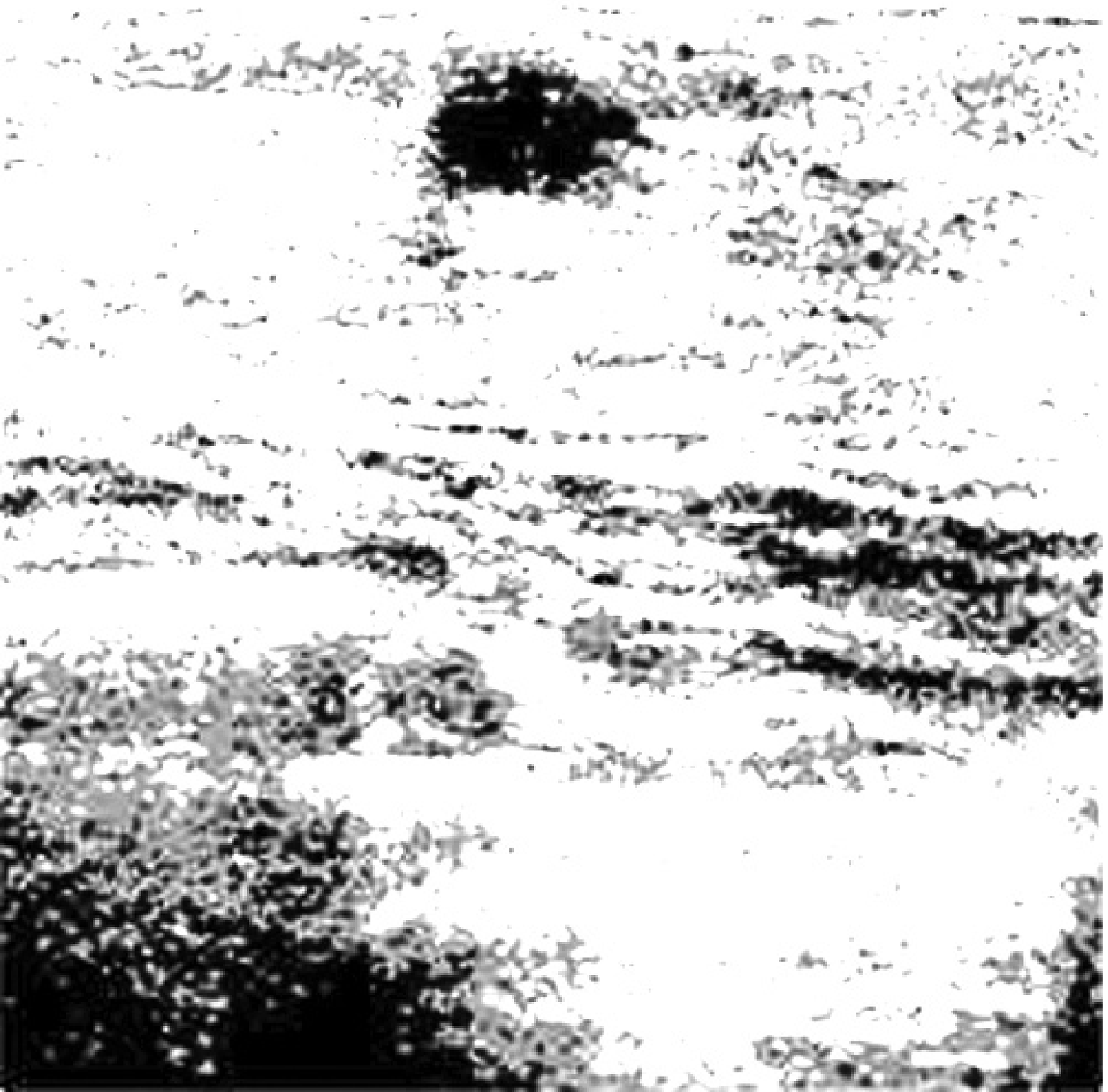} \\
\hbox{\hspace{1.9in}(a) \hspace{1.3in} (b) \hspace{1.3in} (c)}
\includegraphics[width= 1.45in]{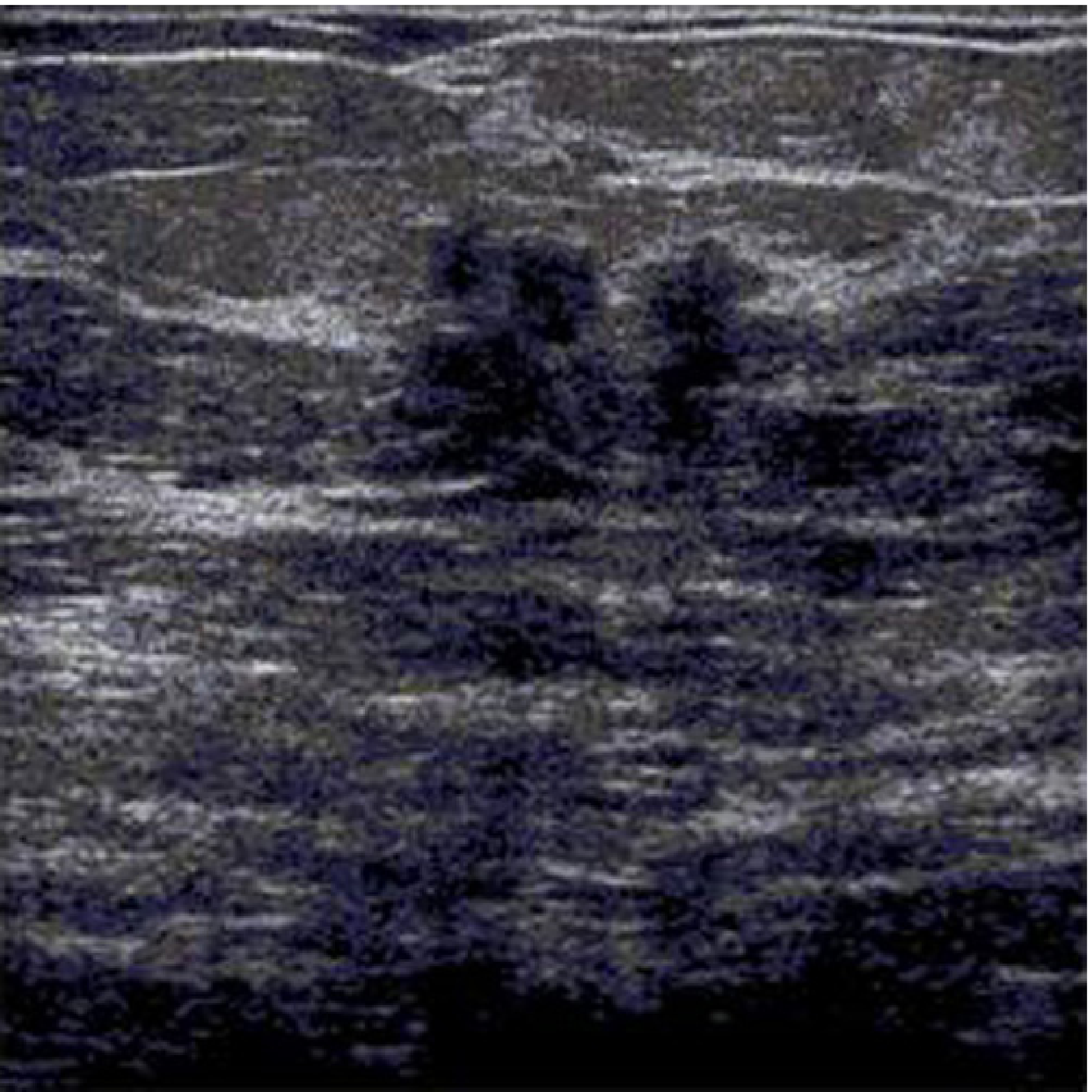}
\includegraphics[width= 1.45 in]{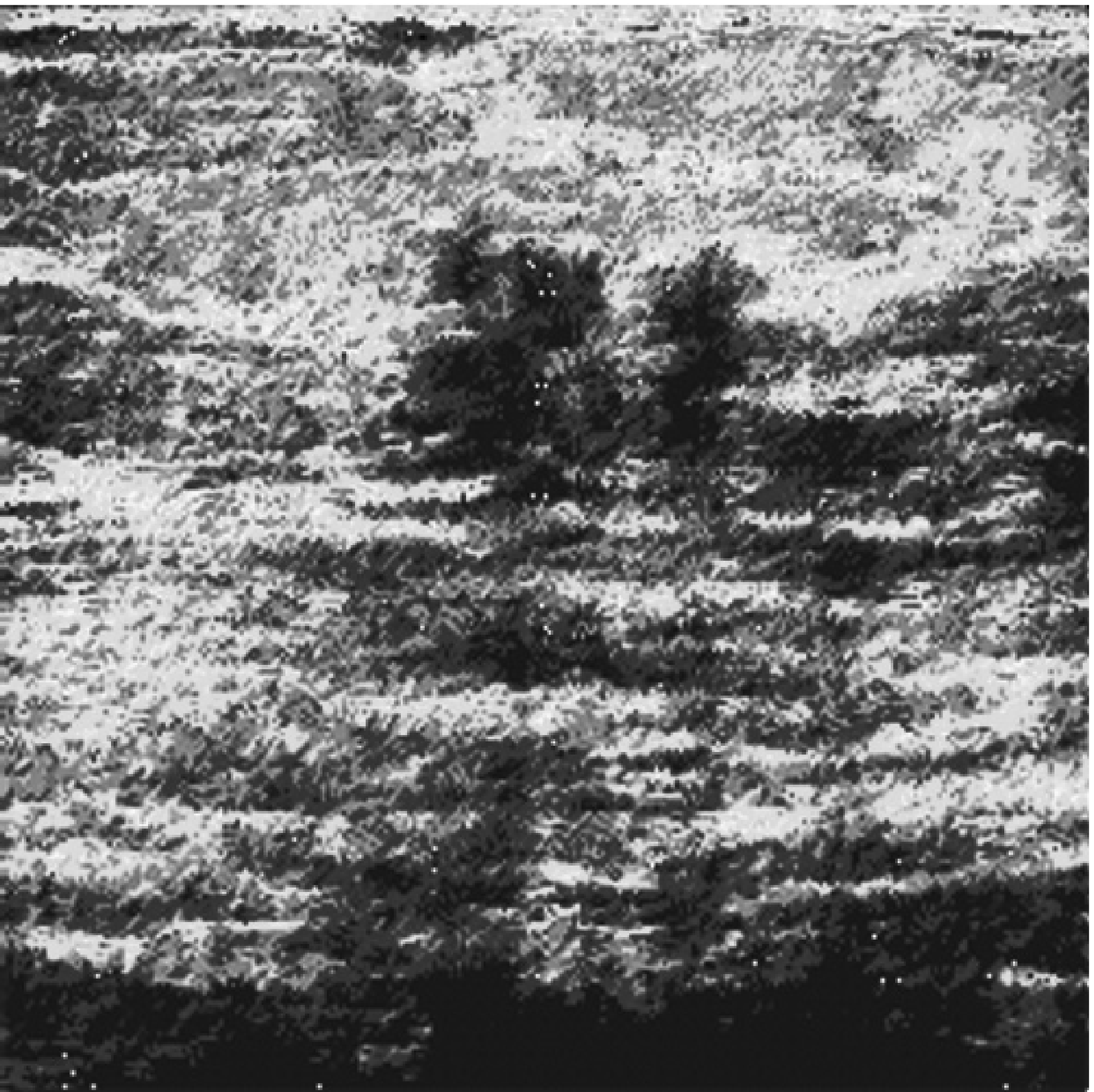}
\includegraphics[width= 1.45 in]{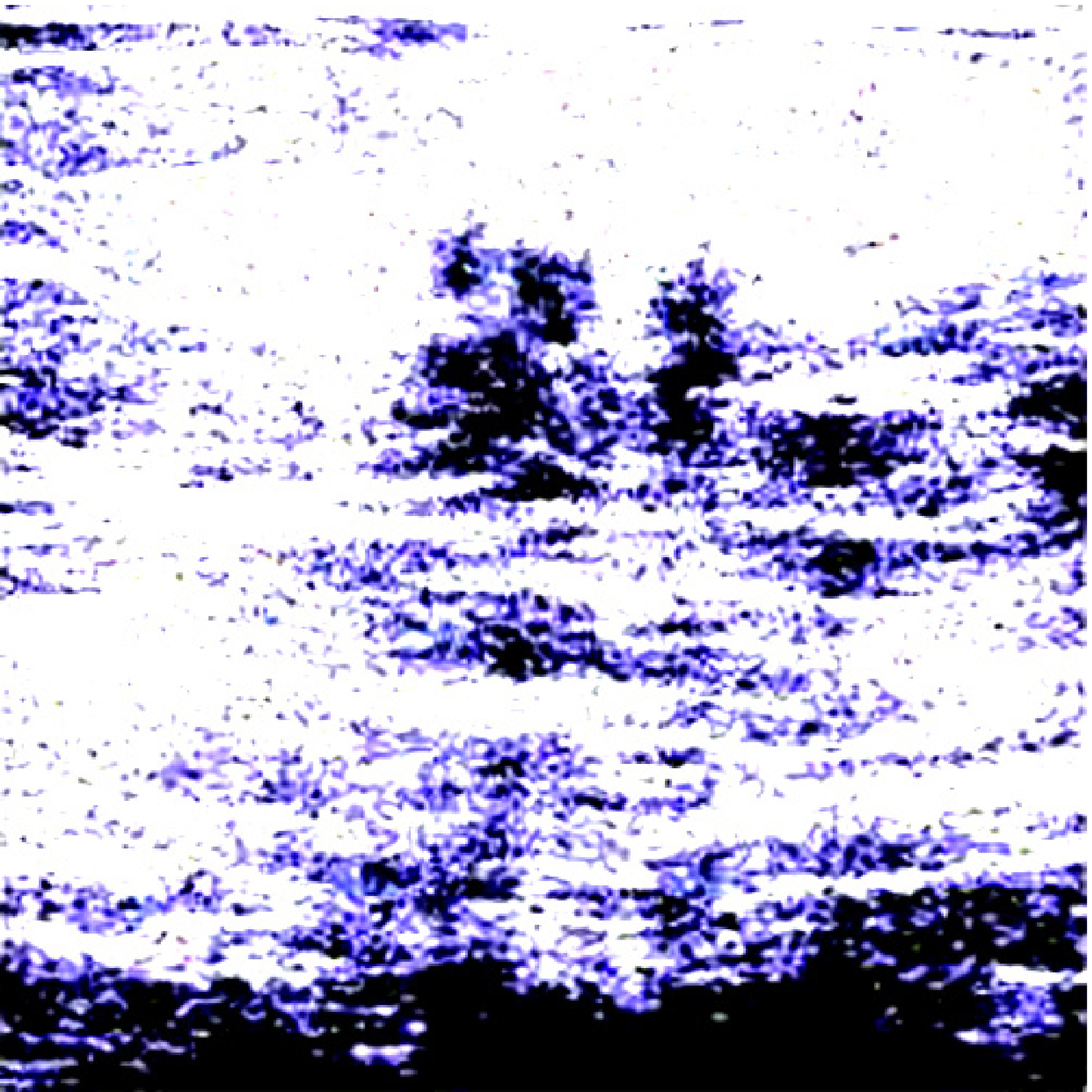} \\
\hbox{\hspace{1.9in}(d) \hspace{1.3in} (e) \hspace{1.3in} (f)}
  \caption{ARMA modeling and segmentation of ultrasound breast images: (a) cancerous ultrasound image; (b) ARMA representation of (a);
(c) Segmentation of (a); (d) benign tumor ultrasound image; (e) ARMA
representation of (d); (f) Segmentation of (d). }\label{fig1}
\end{figure*}

\section{Simulations} \label{sec:3}
Although the proposed algorithm is independent of the imaging
modality of the breast, we perform our simulations on ultrasound
images, collected from the Radiology department, College of Medicine
at the University of Illinois at Chicago. Our database of cancerous
images show intraductal carcinoma, which is the most common type of
breast cancer in women. Intraductal carcinoma is usually discovered
through a mammogram or an ultrasound as microcalcifications. Our
benign tumor images show the Fibroadenoma of the breast, which is a
benign fibroepithelial tumor characterized by proliferation of both
glandular and stromal elements.

We estimate the ARMA parameters using a window of size $16 \times
16$. The choice of the window size presents an inherent trade-off
between the accuracy of the representation and the accuracy of the
classification. A large window size would lead a better
representation of the ARMA model, but might include pixels from
different classes. We found that for $256 \times 256$ images, a $16
\times 16$ window size leads to a good segmentation performance.
Figure \ref{fig1} shows a cancerous image and a benign tumor image.
Their ARMA representations are displayed in Figs. \ref{fig1}(b) and
\ref{fig1}(e), respectively. It is visually clear that the ARMA
model accurately represents both ultrasound images. Figures
\ref{fig1}(c) and \ref{fig1}(f) show the segmentation outputs of the
cancerous and benign tumor images, respectively. We can observe
clear delineations of the tumors from the healthy tissues in both
cases. Using the University of Illinois at Chicago's database of
ultrasound breast images, our method yields an accuracy (number of
images correctly classified divided by the total number of images)
of 82\% (see Table \ref{table1}).

\begin{table}[h]
\caption{Detection accuracy for the University of Illinois at
Chicago's database of ultrasound breast images}\label{table1}
  \centering
\begin{tabular}{|c|c|}
\hline
Accuracy & $82 \%$ \\
\hline
\end{tabular}
\end{table}

\section{Conclusion} \label{sec:conclusion}
We showed that breast images can be accurately represented by 2D
ARMA models. Furthermore, the parameters of the model can be thought
of as the fingerprints of the image, which are exploited for tumor
detection and classification. Our simulation results show that,
based on the estimated ARMA parameters as feature vectors, the
classic k-means algorithm genuinely segment the breast images into
healthy tissue, cancerous tumor and benign tumor. The end product is
a computer-aided diagnosis (CAD) system that clinically portends an
accurate prognosis of breast cancer.

\section {Acknowledgements}
The authors want to thank Dr. Karen Lin Xie from the Radiology
department at the University of Illinois at Chicago for providing
the ultrasound breast images.

%Although computer-aided mammography has been studied over the last
%two decades, automated interpretation of microcalcifications still
%remains very difficult. The major reasons are as follows. First, the
%objects of interest can be extremely small, leading to potential
%misidentification. Second, different sizes, various shapes, and
%variable distributions of microcalcifications appear in mammograms,
%therefore, sample matching seems to be impossible. Third, the
%regions of interest (ROI's) may be of low contrast. The intensity
%difference between suspicious areas and their surrounding tissues
%can be quite slim. Fourth, the dense tissues and/or skin thickening,
%especially in younger women, cause suspicious areas to be almost
%invisible. Finally, dense tissues may be easily misinterpreted as
%calcifications yielding a high false-positive (FP) rate that is a
%major problem with most of the algorithms.

%\bibliographystyle{IEEEbib}
%\bibliography{refs}

\end{document}